\documentclass{article}

\PassOptionsToPackage{compress}{natbib}
\usepackage[nonatbib,final]{nips_2016}

\usepackage[utf8]{inputenc} 
\usepackage[T1]{fontenc}    
\usepackage{hyperref}       
\usepackage{url}            
\usepackage{booktabs}       
\usepackage{amsfonts}       
\usepackage{nicefrac}       
\usepackage{microtype}      
\usepackage{xcolor}

\usepackage{epsfig}
\usepackage{graphicx}
\usepackage{amsmath}
\usepackage{amssymb}
\usepackage{graphics}
\usepackage{subfigure}

\usepackage{algorithm}
\usepackage{algorithmic}
\title{Localizing by Describing: Attribute-Guided Attention Localization for Fine-Grained Recognition}

\author{
  Xiao Liu\footnotemark[1] \hspace{0.15in}~Jiang Wang\thanks{Equal contribution.}\hspace{0.15in} ~Shilei Wen\hspace{0.15in} ~Errui Ding \hspace{0.15in}  ~Yuanqing Lin\\
  Baidu Research\\
  \texttt{\{liuxiao12, wangjiang03, wenshilei, dingerrui, linyuanqing\}@baidu.com} \\
}

\begin{document}

\maketitle

\setlength{\textfloatsep}{8pt}

\begin{abstract}
A key challenge in fine-grained recognition is how to find and represent discriminative local regions.
Recent attention models are capable of learning discriminative region localizers only from category labels with reinforcement learning. However, not utilizing any explicit part information, they are not able to accurately find multiple distinctive regions.
In this work, we introduce an attribute-guided attention localization scheme where the local region localizers are learned under the guidance of part attribute descriptions.
By designing a novel reward strategy, we are able to learn to locate regions that are spatially and semantically distinctive with reinforcement learning algorithm. The attribute labeling requirement of the scheme is more amenable than the accurate part location annotation required by traditional part-based fine-grained recognition methods.
Experimental results on the CUB-200-2011 dataset~\cite{bd6} demonstrate the superiority of the proposed scheme on both fine-grained recognition and attribute recognition.
\end{abstract}

\section{Introduction}
\begin{figure}[!t]
\begin{center}
    \subfigure[Two types of birds with part-attribute descriptions.]{
    \includegraphics[scale = 0.4]{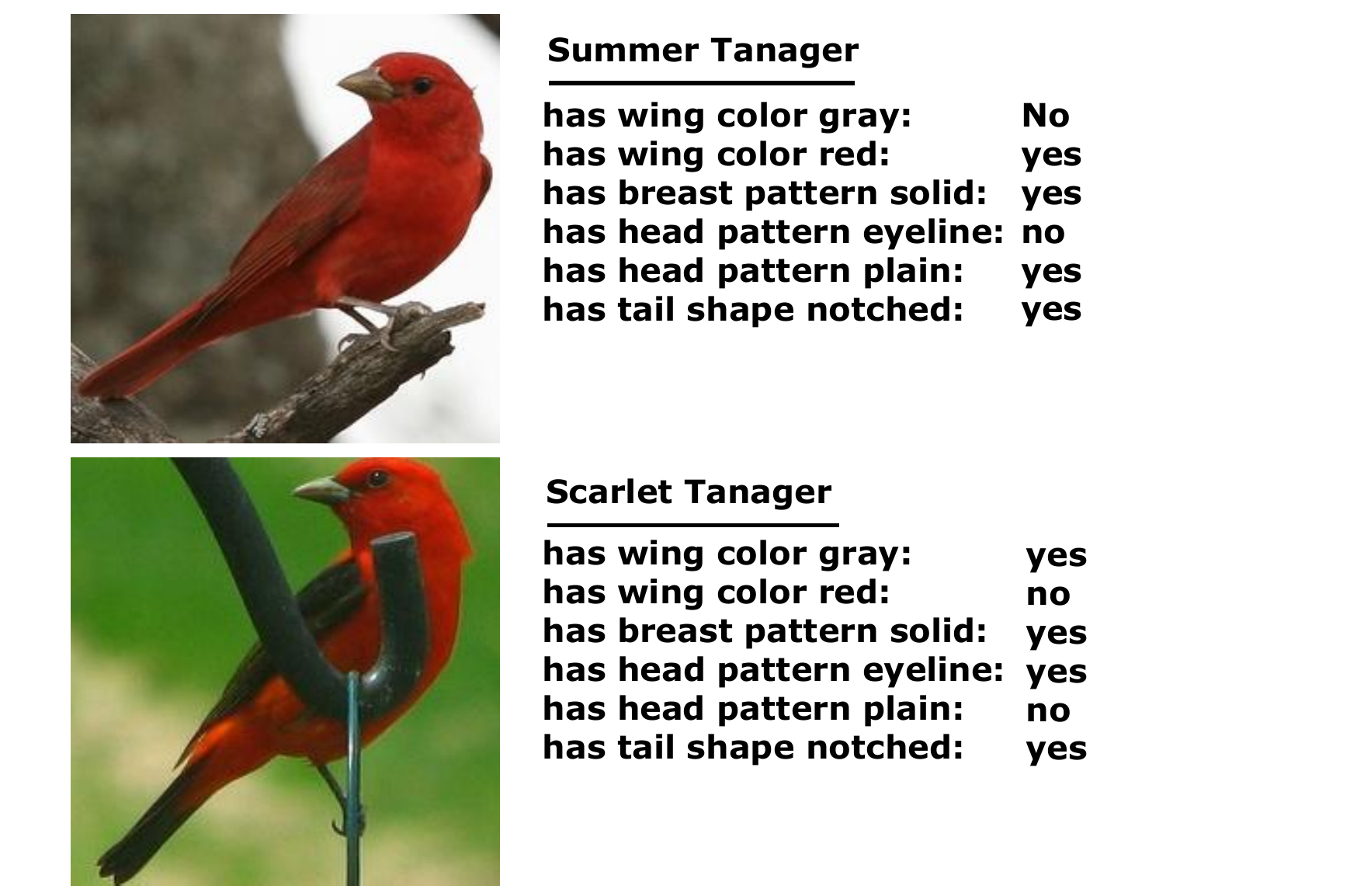}}
    \hspace{0.1in}
    \begin{minipage}[b]{5cm}
    \centering
    \subfigure[Attention regions found by \cite{nips1}.]{
    \includegraphics[scale = 0.35]{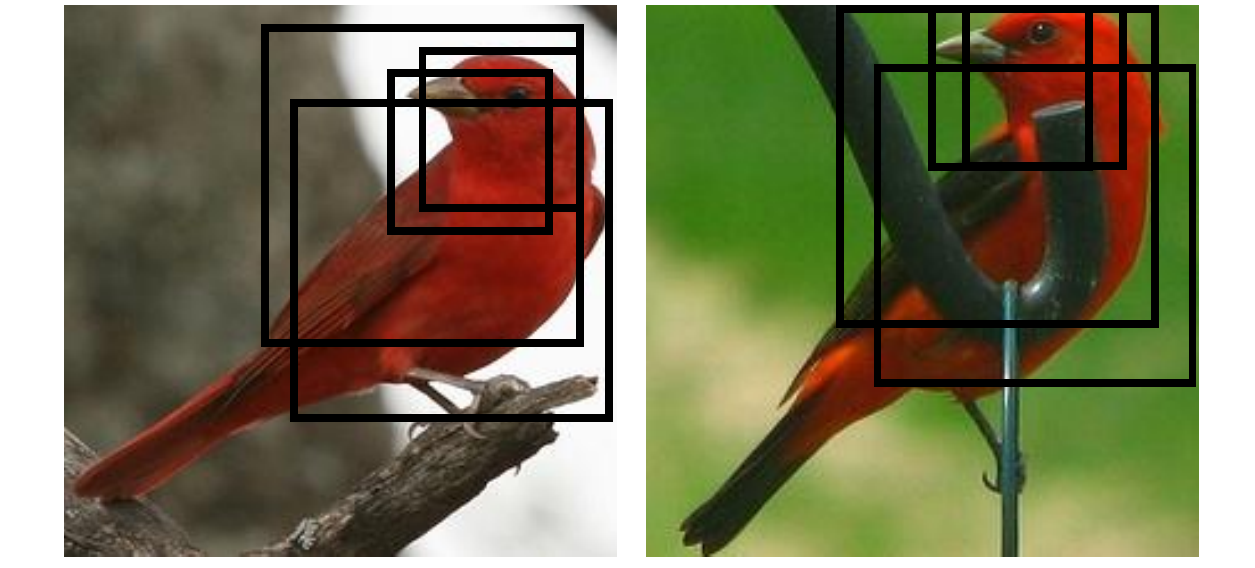}
    }\\
    \subfigure[Part-attribute-guided attention. ]{
    \includegraphics[scale = 0.35]{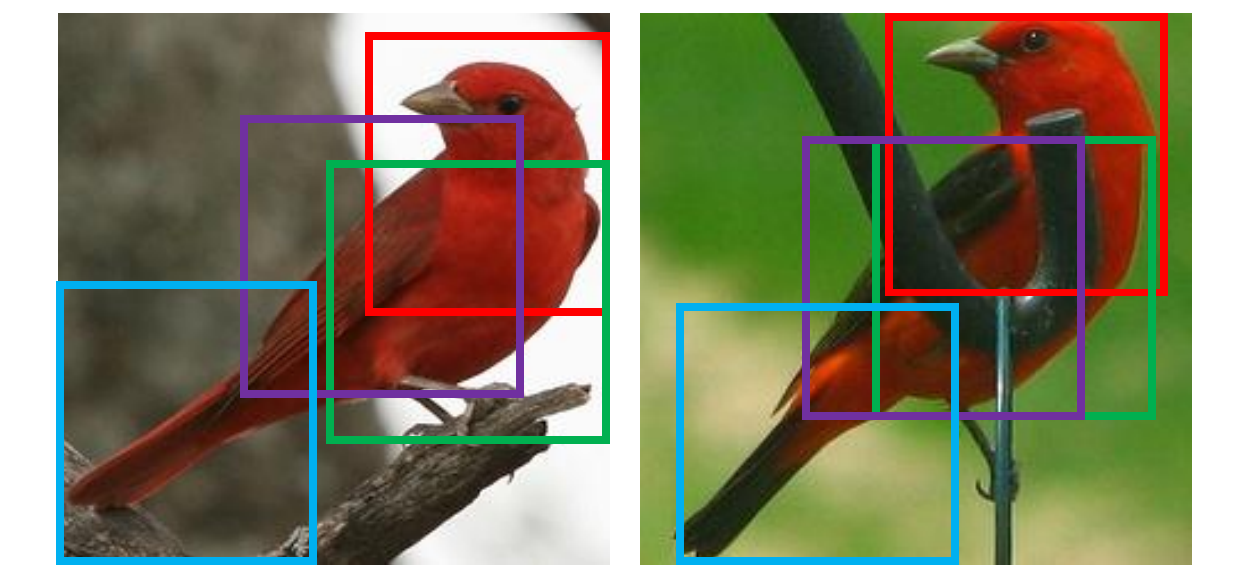}
    }
    \end{minipage}
\vspace{-12pt}
\end{center}
\label{fig:fig0}
\caption{(a) Two types of birds with part-attribute descriptions. Human experts differentiate a summer tanager and a scarlet tanager by the color of wing and tail. (b) \cite{nips1} fails to locate the tail part.
(c) Part-attribute-guided attention localization networks can locate object parts more accurately. The red, green, purple and blue bounding boxes localize head, breast, wing and tail, respectively (best viewed in color). }
\end{figure}

Humans heavily rely on subtle local visual cues to distinguish fine-grained object categories.
For example in Figure~\ref{fig:fig0} (a), human experts differentiate a summer tanager and a scarlet tanager by the color of wing and tail.  In order to build human-level fine-grained recognition AI systems, it is also essential to locate discriminative object parts and learn local visual representation for these parts.

State-of-the-art fine-grained recognition methods \cite{bd11,bd13} either rely on manually labeled parts to train part detectors in a fully supervised manner, or employ reinforcement learning or spatial-transformer-based attention models \cite{bd3,nips1} to locate object parts with object category annotations in a weakly supervised manner. However, both types of methods have major practical limitations.
Fully supervised methods need time-consuming, error-probing manual object part labeling process, while object labels alone as a supervision signal is generally too weak to reliably locate multiple discriminative object parts.
For example in Figure~\ref{fig:fig0}(b), the method~\cite{nips1} fails to locate the tail as attention regions.

Humans have a remarkable ability to learn to locate object parts from multiple sources of information. Aside from strongly labeled object part location and weakly labeled object categories, part descriptions,  such as ``red wing'', also plays an important part in the development of object parts locating ability. In their early ages, children learn to recognize parts and locate object parts by reading or listening to part descriptions. Part descriptions do not require time-consuming manually part location labeling, and it is much stronger than object category labels. We call it {\em part attribute}.

Inspired by this capability, we propose a part attribute-guided attention localization scheme for fine-grained recognition.
Using part attributes as a weak supervision training signal, reinforcement learning is able to learn part-specific optimal localization strategies given the same image as environment state.
Based on this intuition, it is reasonable to expect that distinctive part localizers could be learned as strategies of  looking for and describing the appearances of different parts.
In the proposed scheme, multiple fully convolutional attention localization networks~\cite{nips1} are trained. Each network predicts the attribute values of a part. We design a novel reward strategy for learning part localizers and part attribute predictors.

Part attribute-guided attention localization networks can more accurately locate object parts (Figure~\ref{fig:fig0}(c)). More importantly, using the part locations and appearance features from part-attribute guided attention localization networks leads to state-of-the-art performance on fine-grained recognition, as demonstrated on the CUB-200-2011 dataset~\cite{bd6}.
Moreover, part attribute can be acquired in large scale via either human labeling or data mining techniques. It has been successfully employed for image recognition~\cite{parikh2011relative,akata2013label,hwang2014unified}, image retrieval~\cite{huang2015cross}, and image generation~\cite{yan2015attribute2image}.  To our best knowledge, no one has utilized part attribute to guide the learning of visual attention.


\section{Attribute-Guided Attention Localization}
\begin{figure}[!t]
\begin{center}
\includegraphics[scale = 0.3]{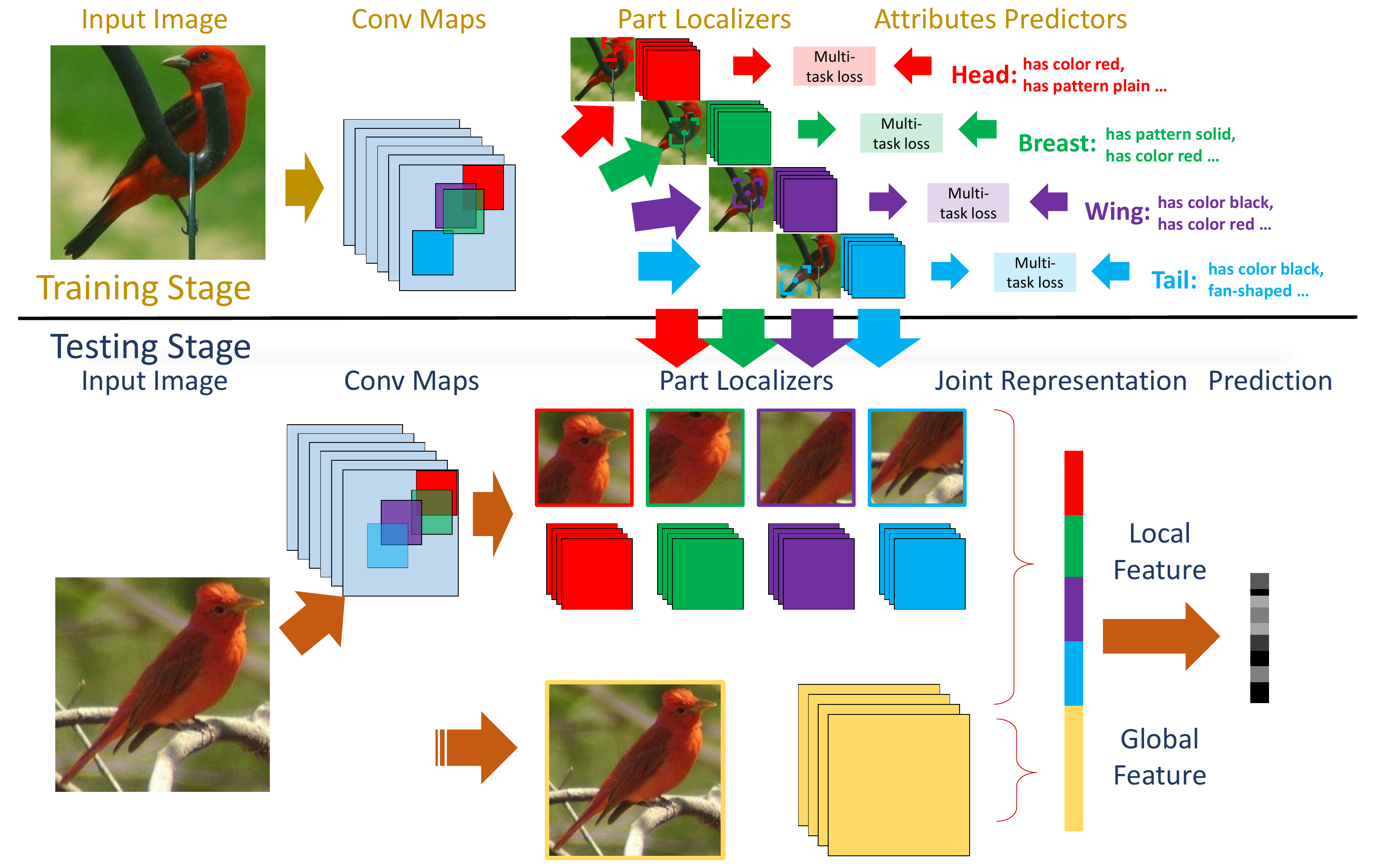}
\end{center}
\caption{An overview architecture of the proposed scheme. The correspondence of color and part is the same as Fig.~\ref{fig:fig0}.
The upper part shows training stage, and the lower part shows the testing stage. In the training stage, multiple part localizers are trained under the guidance of attribute descriptions.
In the testing stage, features extracted from the selected part regions and the entire image are combined into a joint representation for category prediction.
}\label{fig:fig1}
\end{figure}

The architecture of the proposed scheme is shown in Figure~\ref{fig:fig1}.
Attribute descriptions are only used for learning part localizers during training. They are not used in the testing stage.

In the training stage, a fully-convolutional attention localization network \cite{nips1} is learned for each part.
In contrast with \cite{nips1}, the task of the fully-convolutional attention localization network is to learn where to look for better part attribute prediction rather than predicting the category of the entire object.

For testing, we extract features from both part regions located by the network and the entire image and concatenate them as a joint representation. The joint representation is utilized to predict the image category.

\subsection{Problem Formulation}
Given $N$ training images $\{x_1, x_2, \dots  , x_N\}$, and their object labels $\{l_1, l_2, \dots, l_N \}$. Our goal is to train a model for classifying each image $x_i$ as its ground-truth label $l_i$.

Fine-grained objects have $P$ parts. \emph{Localizing by describing} problem finds a policy to locate the parts and to classify images based on these part locations and their features. The objective can be formulated as:

\begin{equation}\label{eq:loss1}
\mathcal{L}_0  = \sum_i L(\mathcal{G}(x_i, \mathcal{F}_1(x_i), \dots, \mathcal{F}_P(x_i)), l_i),
\end{equation}

where $\mathcal{F}_p(\cdot)$ is a function that finds the location of part $p$ and crops its image region.
The crop size of each part is manually defined.
$\mathcal{G}(\cdot)$ is a deep convolutional neural network classifier that outputs the probability of each category given the whole image and the cropped image regions for all the parts.
$L(.)$ is cross-entropy loss function measuring the quality of the classification.

Precisely predicting part locations using only the image labels is very challenging.
In {\em localizing by describing}, we localize object parts with the help of visual attribute descriptions.
The visual attribute is a set of binary annotations whose value correlates with an aspect of the object appearance.
A local attribute is generally related to the appearance of an object part.
The attribute description of the $p$-th part is denoted as $\{Y^p_1, Y^p_2, \dots  , Y^p_N\}$.
Each part description $Y^p_i$ is a binary vector: $[y^p_{i,1},y^p_{i,2},\dots,y^p_{i,K_p}]$, where each element $y^p_{i,k}$ indicates whether an attribute exists in the $i$-th image, and $K_p$ is the number of attributes for the $p$-th part.
We aim to learn better part localizers $\mathcal{F}_p$ using part attribute annotation.
An auxiliary part attribute classification loss is proposed to facilitate the learning of part localizers
\begin{equation}
\mathcal{L}_p  = \sum_i L'(\mathcal{T}_p(\mathcal{F}_p(x_i)), Y^p_i),
\end{equation}
where $\mathcal{T}_p(.) = [\mathcal{T}_{p,1}(.),\dots,\mathcal{T}_{p,K_p}(.)]$ is a multi-label attribute prediction function of the $p$-th part, and each $\mathcal{T}_{p,k}(.)$ indicates the predicted probability of the $k$-th attribute of part $p$.
$L'(.)$ is a multi-label cross-entropy loss
\begin{equation}\label{lab:multi_cross_entropy}
L'(X_i, Y^p_i) = -\sum_{k} \left[y^p_{i,k} \log \mathcal{T}_{p,k}(\mathcal{F}_p(x_i)) + (1-y^p_{i,k})\log(1- \mathcal{T}_{p,k}(\mathcal{F}_p(x_i)))\right].
\end{equation}

The assumption is that the part localizers that help predict part attributes are also beneficial to the prediction of the whole object.
We use Eq.~\eqref{lab:multi_cross_entropy} as an auxiliary loss to learn localizer $\mathcal{F}_p(\cdot)$ that optimizes Eq.\eqref{eq:loss1}.

\subsection{Training of Localizers}
Given images and attribute descriptions, we jointly learn a part localizer and a multi-label predictor for each part such that the predictor uses the selected local region for attribute prediction.

Since the localization operation is non-differential, we employ reinforcement learning algorithm \cite{bd20} to learn the part localizers and multi-label predictors.
For reinforcement learning algorithm, the policy function is the part localizer function $\mathcal{F}_p(\cdot)$;
the state is the cropped local image patch $\mathcal{F}_p(x_i)$ and the whole image $x_i$;
the reward function measures the quality of the part attribute and object prediction.

The objective function of the reinforcement learning algorithm for part $p$ is
\begin{equation}\label{eq:reinforcement}
J_p(\mathcal{F}_p, \mathcal{T}_p) = \sum_i  \left[  \mathbb{E}_{\mathcal{F}_p}(R_{p,i}) - \lambda  L'(\mathcal{T}_p(\mathcal{F}_p(x_i)), Y^p_i) \right ],
\end{equation}
where the reward
\begin{equation}
\mathbb{E}_{\mathcal{F}_p}(R_{p, i}) = \int_{s_{p,i}} p_{\mathcal{F}_p} (s_{p,i}|x_i)r(s_{p,i})d(s_{p,i})
\end{equation}
is the expected reward of the selected region for the $p$-th part of the $i$-th image.
$s_{p,i}$ indicates a selected region, $p_{\mathcal{F}_j} (s_{p,i}|x_i)$ is the probability that the localizer select the region $s_{p,i}$.
$r(s_{p,i})$ is a reward function to evaluate the contribution of the selected region $s_{p,i}$ to attribute prediction.

Previous methods~\cite{bd3,nips1} choose the reward function $r(s_{p,i})$ to be $1$ only when the image is correctly classified.
However, since our algorithm predicts multiple attribute values for a part, it is too strict to enforce all the attributes are correctly predicted.
Therefore, we consider an alternative reward strategy.
A selected region has reward 1 if both of the following criteria are satisfied:
1) it achieves lower attribute classification loss than most  other regions in the same image,
\emph{i.e}., its prediction loss ranks top-$\eta$ lowest among the $M$ sampled regions of the image.
2) it achieves lower attribute classification loss than most other regions in the same mini-batch,
\emph{i.e}., its prediction loss is lower than half of the average loss of all the regions in the mini-batch.

Following \cite{nips1}, we learn fully convolutional attention localization networks as part localizers.
Since both parts of the objective function Eq.~\eqref{eq:reinforcement} are differentiable,
REINFORCE algorithm~\cite{bd20} is applied to compute the policy gradient to optimize the objective function:
\begin{eqnarray}
\partial_{\mathcal{F}_p} \mathbb{E}_{\mathcal{F}_p}(R_{p,i}) \hspace{2.5in}\\
= \int_{s_{p,i}} p_{\mathcal{F}_p} (s_{p,i}|x_i)   \partial_{\mathcal{F}_p} \log   [p_{\mathcal{F}_p}      (s_{p,i}|x_j)r(s_{p,i})] d(s_{p,i})  \nonumber \\
\approx \frac{1}{M} \sum_{m=1}^M  \partial_{\mathcal{F}_p} \log   [p_{\mathcal{F}_p}      (s^m_{p,i}|x_i)r(s^m_{p,i})]  \hspace{0.9in}  \nonumber
\end{eqnarray}
where $s^m_{i,j} \sim p_{\mathcal{F}_i} (\cdot|x_j)$ is the local image regions sampled according to localizer policy $p_{\mathcal{F}_i} (s^m_{i,j}|x_j)$.
$M$ local regions of the same image are sampled in a mini-batch.
We list the learning algorithm in Algorithm 1.

\begin{algorithm}[t]
\caption{Learning algorithm for localizing by describing: }
\begin{algorithmic}[1]
\renewcommand{\algorithmicrequire}{\textbf{Input:}}
\renewcommand{\algorithmicensure}{\textbf{Output:}}
\REQUIRE{ training images $\{x_1, x_2, \dots  , x_N\}$, attribute descriptions of each part $\{Y^p_1, Y^p_2, \dots  , Y^p_N\}$.}
\ENSURE{part localization function $\mathcal{F}$, multi-label attribute prediction function $\mathcal{T}$.}

\FOR{each part $p$}

\STATE {Initialize $\mathcal{F}_p$ and $\mathcal{T}_p$.}

\REPEAT

\STATE Randomly sample $H$ images.

\FOR{each image $x_i$}

\STATE Sample $M$ regions according to the output of the part localizer: $s^m_{p,i} \sim p_{\mathcal{F}_p} (\cdot|x_i)$

\ENDFOR

\FOR{each local region $s^m_{p,i}$}

\STATE Calculate the multi-label attribute prediction loss $l_{p,i}^m = L'(\mathcal{T}_p(s^m_{p,i}), Y^p_i)$.

\ENDFOR

\STATE Calculate the average loss of the mini-batch $\hat{L} = \sum l_{p,i}^m$.

\FOR{each image $x_i$}

\STATE Sort $l_{p,i}^m$ in ascending order.

\FOR{each local region $s^m_{p,i}$}

\IF {$l_{p,i}^m$ is in the top-$\eta$ position of the sorted list, and $l_{p,i}^m < 0.5\hat{L}$}

\STATE Set reward $r(s^m_{p,i}) = 1$.

\ELSE

\STATE Set reward $r(s^m_{p,i}) = 0$.

\ENDIF

\ENDFOR

\ENDFOR

\STATE Calculate the policy gradient of $\mathcal{F}_p$ according to (6).

\STATE Calculate the gradient of $\mathcal{T}_p$ by standard back-propagation.

\STATE Update the parameters of $\mathcal{F}_p$ and $\mathcal{T}_p$.

\UNTIL{converge}

\ENDFOR

\end{algorithmic}
\end{algorithm}

\subsection{Training of Classifiers}
After the local region localizers are trained, we re-trained the attribute prediction models using up-scaled local regions.
When re-trained, the attribute predictors takes up-scaled local regions from the part localizers to predict the attributes.

To combine global and local information, we extract features from all the part regions and the entire image and concatenate them to form a joint representation. The joint representation is used to predict image category.
In details, we first train classifiers with each individual part region to capture the appearance details of local parts for fine-grained recognition.
A classifier utilizing the entire image is also trained for global information.
We then concatenate features extracted from all the parts and the entire image as a joint representation, and we use a linear layer to combine the features.

\subsection{Prediction}
The prediction process is illustrated in the lower part of Fig.~\ref{fig:fig1}.
We localize and crop each part using its localizer $\mathcal{F}_p(x_i)$,
and each part region is resized to high resolution for feature extraction.
Features from all part regions as well as the entire image are concatenated as a joint representation. A linear classification  layer is utilized to make the final category prediction.

We also attempt to use attribute prediction results to help fine-grained object recognition, or model the geometric relationship of the parts using recurrent convolutional operation.
However, we find neither of the approaches achieve notable improvements in our experiments.
Detailed experimental results and setup can be found in the experimental section.

\section{Experiments}
We conduct experiments on the CUB-200-2011 datasets \cite{bd6}. The dataset contains $11,788$ images of $200$ bird categories, where $5,994$ images are for training, and the rest $5,794$ images are for testing.
In addition to the category label, 15 part locations, 312 binary attributes and a tight bounding box of the bird is provided for each image. Examples of images and attributes in this dataset are shown in Figure \ref{fig:fig0}(a).

\subsection{Implementation Details}
We evaluate the proposed scheme in two scenarios: ``with BB'' where the object bounding box is utilized during training and testing,
and ``without BB'' where the object bounding-box is not utilized.

We choose 4 parts, \emph{i.e.} ``head'', ``wing'', ``breast'', and ``tail'', to train local region localizers.
The cropping size of these parts are half of the original image.
Among all the 312 attributes, if a part name appears in an attribute, then the attribute is considered describing this part.
The number of attributes describing the four parts are 29, 24, 19, and 40, respectively.

We utilize ResNet-50~\cite{nips2} as the visual representation for part localization and feature extraction.
In the training stage, we utilize the entire image (in the ``with BB'' setting, crop the image region within the bounding box) to learn multi-label attribute predictions for each part.
The output of the ``res5c'' layer of ResNet-50 is employed as the input of the fully convolutional attention localization networks. The attribute predictors use ROI-pooled feature maps \cite{girshick2015fast} of the fully convolutional attention localizers.

We train the models using Stochastic Gradient Descent (SGD) with momentum of 0.9, epoch number of 150, weight decay of 0.001, and a mini-batch size of 28 on four K40 GPUs.
One epoch means all training samples are passed through once.
An additional dropout layer with an ratio of 0.5 is added after ``res5c'', and the size of ``fc15'' is changed from 1000 to 200.

The parameters before ``res5c'' are initialized by the model \cite{nips2} pretrained on the ImageNet dataset \cite{bd18},
and parameters of fc15 are randomly initialized.
The initial learning rate is set at 0.0001 and reduced twice with a ratio of 0.1 after 50 and 100 epoches.
The learning rate of the last layer (``fc15'') is 10 times larger than other layers.

Our data augmentation is similar to \cite{bd7}, but we have more types of data augmentation.
A training image is first rotated with a random angle between $-30^\circ$ and $30^\circ$.
A cropping is then applied on the rotated image. The size of the cropped patch is chosen randomly between $25\%$ and $100\%$ of the whole image, and its aspect ratio is chosen randomly between $3/4$ and $4/3$.
AlexNet-style color augmentation \cite{nips3} is also applied followed by random flip. We finally resize the transformed cropped patch to a $448\times448$ image as the input of the convolutional neural network.

\subsection{Part Localization Results}
We report our part localization results in Table 1. Percent Correct Parts (PCP) is used as the evaluation metric.
A part is determined to be correctly localized if the difference of its predicted location and the ground-truth location is within 1.5 times the ground-truth annotation’s standard deviation.

We compare with previous state-of-the-art part localization methods \cite{nips5,nips6,nips7}.
The strongly supervised method \cite{nips5}, which utilizes the precise location of the parts during training,  achieves the highest average PCP (71.4) in the ``without BB'' scenario.
Our scheme that does not use any part or object locations  achieves the second highest average PCP (66.2) and performs the best for localizing ``breast''.
The parts are localized much more precisely ($66.2 \to 73.4$) when ground-truth bird bounding boxes is used.

\begin{figure}[!t]
\begin{center}
\includegraphics[scale = 0.45]{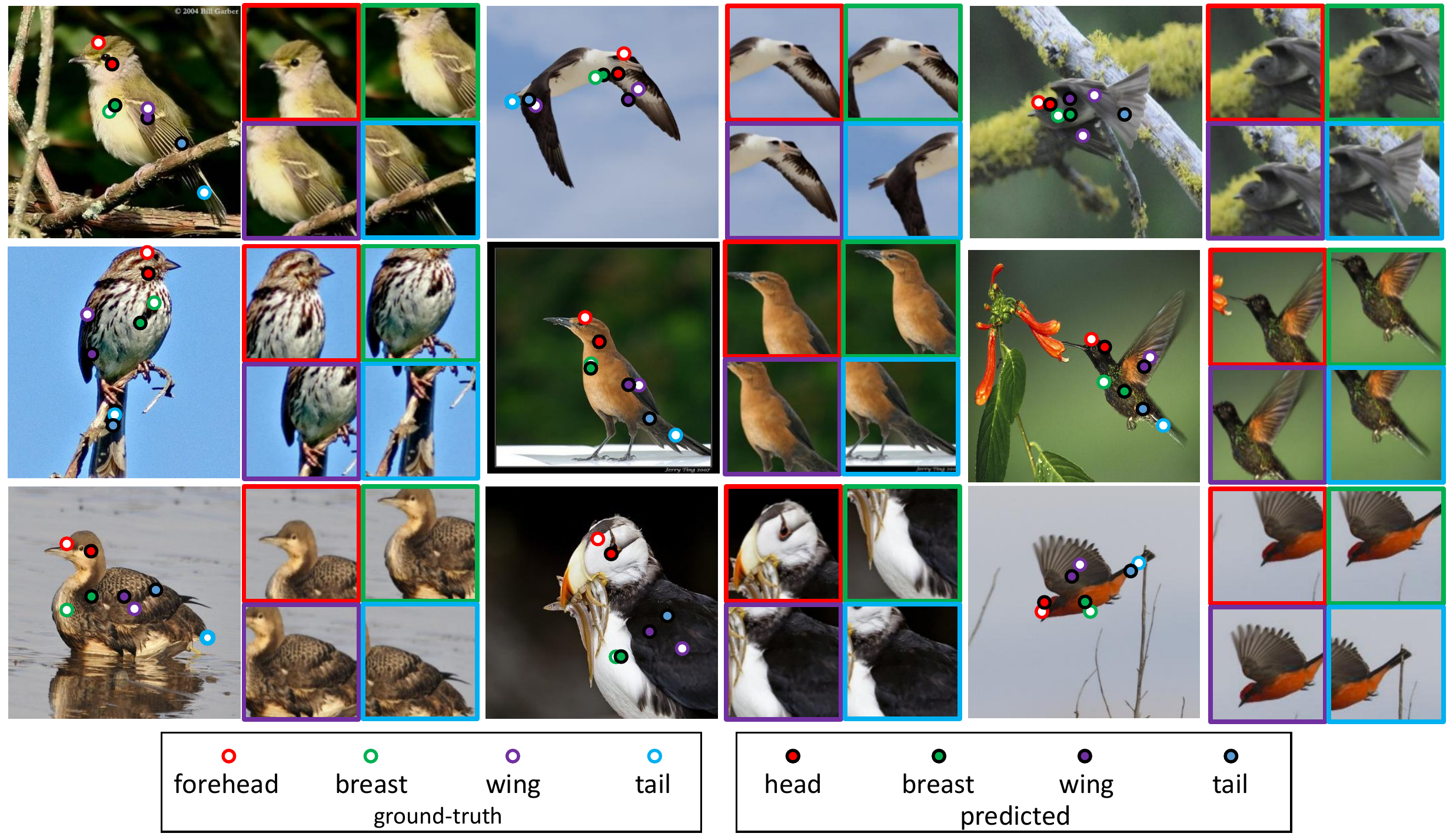}
\end{center}
\caption{Visualizations of part localization results. Ground-truth part locations are shown as hollow circles; predicted part locations are shown as solid circles; selected part regions are shown as thumbnails.
Different parts are shown in different colors. No ground-truth bird bounding boxes or part locations information are used during training or testing time in the proposed method.
}\label{fig:fig3}
\end{figure}

Figure~\ref{fig:fig3} provides visualizations of our part localization results.
Ground-truth part locations are shown as hollow circles, predicted part locations are shown as solid circles and the selected part regions are shown as thumbnails.

It should be noted that different from \cite{nips5}, our scheme does not directly minimize the part localization error but learns part localizers for attribute prediction.
Thus, a selected region that is far from manually annotation might better predict some part attributes.
For example, our predicted ``head'' positions are  usually in the center of the bird head such that the cropped local region does not lose any information,
while the manually annotated ``head'' positions normally appear on the ``forehead''.

\begin{table}[tbh!]
\begin{small}
\begin{center}
\caption{Part localization results (measured by PCP) on the CUB-200-2011 dataset.}
\begin{tabular}
{l||c|c|c|c|c}\hline
Method & head & breast & wing & tail & ave\\\hline
Shih et al. \cite{nips5} & 67.6 & 77.8 & \bf{81.3} & \bf{59.2} & \bf{71.4} \\
Liu et al. \cite{nips6} & 58.5 & 67.0 & 71.6 & 40.2 & 59.3\\
Liu et al. \cite{nips7} & \bf{72.0} & 70.5 & 74.4 & 46.2 & 65.8\\
Ours (without BB) & 60.6 & \bf{79.5} & 77.5 & 47.1 & 66.2 \\\hline
Ours (with BB) & 69.3 & 81.5 & 80.3& 62.5 & 73.4\\\hline
\end{tabular}
\end{center}
\end{small}
\vspace{-20pt}
\end{table}

\subsection{Attribute Prediction Results}
The attribute prediction results measured by average Area Under the Curve of ROC (AUC) are reported in Table 2.
We use AUC instead of accuracy because the data of attributes are highly imbalanced:
most attributes are only activated in a very small number of images, but some attributes appear very frequently. For each part, we calculate the AUC of all its attributes, and report the average AUC of them as the AUC of the part.

Directly utilizing the full image as input achieves $76.8$, $78.7$, $73.1$, and $70.4$ average AUC for parts ``head'', ``breast'', ``wing'' and ``tail'', respectively.
The overall average AUC of the four parts is 74.3.
Using the localized local regions results in slight performance drop ($74.3\to 73.7$) on overall average AUC.
The prediction using both the full image and the local attention regions improves the overall average AUC result to 76.6.
Bounding boxes of birds are not used for attribute prediction.

\begin{table}[tbh!]
\begin{small}
\begin{center}
\caption{Attribute prediction results (measured by average AUC) on the CUB-200-2011 dataset.}
\begin{tabular}
{l||c|c|c|c|c}\hline
Method & head & breast & wing & tail & total\\\hline
Full image & 76.8 & 78.7 & 73.1 & 70.4 &  74.3 \\
Attention  & 77.8 &  78.3 & 71.9 & 68.8 & 73.7 \\
Full Image + Attention & \bf{80.3} &  \bf{80.7} & \bf{75.5} & \bf{72.1} & \bf{76.6} \\\hline
\end{tabular}
\end{center}
\end{small}
\end{table}

\subsection{Fine-grained Recognition Results}
For ``without BB'' scenario, the baseline ResNet-50 using the whole image achieves $81.7\%$ recognition accuracy.
Adding features of two parts (``head'' and ``breast'') improves the result to $84.7\%$ and combing features of four parts improves the result to $85.1\%$.

For ``with BB'' scenario, the baseline achieves $82.3\%$ accuracy.
Combing features of two parts improves the result to $84.9\%$, and combing all the features improves the accuracy to $85.3\%$.

We carry out further experiments to explore using the attributes for better recognition.

In the ``full image + attribute value'' experiment, we concatenate 112 binary attribute labels with the original visual  feature of the whole image to predict the bird category.
In the ``full image + attribute feature'' experiment, we concatenate the visual features of the 5 part attribute prediction models:
one is the original full image model, and the other four models are fine-tuned for predicting the attributes of the four parts.

As Table 3 shows, directly combining attribute values does not improve recognition accuracy compared with the baseline.
Combing attribute features leads to marginal improvements ($81.7\% \to 82.5\%$ and $82.3\% \to 82.9\%$), because we find the attribute predictions are usually noisy due to the inherent difficulty of predicting some local attributes .
By combining features from the full image model, four part-based models and the attribute prediction models, we achieve the $85.4\%$ and $85.5\%$ for the ''without BB'' and ``with BB'' scenarios, respectively.

We also explore jointly training the part localizers using a recurrent convolutional neural network model as  geometric regularization of part locations \cite{nips8},
but we find the accuracy improvement is negligible ($85.1\% \to 85.2\%$ without BB). After examining the data, we find birds’ poses are too diverse to learn an effective geometric model from limited amount of data.

\begin{table}[thb!]
\begin{small}
\begin{center}
\caption{Recognition results on the CUB-200-2011 dataset with different settings.}
\begin{tabular}
{l||c|c}\hline
Method & Acc without BB(\%) & Acc with BB(\%)\\\hline
Lin et al. \cite{bd16} &       84.1  &     85.1\\
Krause et al. \cite{bd22}  &   82.0 &      82.8\\
Zhang et al. \cite{bd11}  &73.9 &    76.4 \\
Liu et al. \cite{nips1}      & 82.0 &    84.3\\
Jaderberg et al. \cite{jaderberg2015spatial} & 84.1 & - \\
\hline
Full Image & 81.7  & 82.3\\
Full Image + 2$\times$parts & 84.7  & 84.9 \\
Full Image + 4$\times$parts & 85.1  & 85.3 \\
Full Image + attribute value & 81.7  &  82.2\\
Full Image + attribute feature & 82.5 &  82.9\\
Full Image + 4$\times$parts + attribute feature & \bf{85.4}  & \bf{85.5}  \\\hline
\end{tabular}
\end{center}
\end{small}
\end{table}

We compare with previous state-of-the-art methods on this dataset and summarize the recognition results in Table 3.
The proposed scheme outperforms state-of-the-art methods that use the same amount of supervision.

Lin et al. \cite{bd16} construct high dimensional bilinear feature vectors, and achieve $84.1\%$ and $85.1\%$ accuracy for the ''without BB'' and ``with BB'' scenarios, respectively.
Krause et al. \cite{bd22} learn and combine multiple latent parts in a weakly supervised manner, and achieve $82.0\%$ and $82.8\%$ accuracy for the ''without BB'' and ``with BB'' scenarios, respectively.
Zhang et al. \cite{bd11} train part-based R-CNN model to detect the head and body of the bird. The method relies on part location annotation during training.
Our scheme outperforms \cite{bd11} by a large margin without requiring strongly supervised part location annotation.
Liu et al. \cite{nips1} utilize the fully convolutional attention localization networks to select two local parts for model combination. The accuracy is $82.0\%$ without bounding box and $84.3\%$ with bounding box, while our accuracy is $84.7\%$ without bounding box and $84.9\%$ with bounding box by combing features from two local parts (``head'' and ``breast''). Localizing distinctive parts leads to better recognition accuracy.
Similarly, Jaderberg et al. \cite{jaderberg2015spatial} combine features of four local parts and achieve an accuracy of $84.1\%$ without using bounding box. Our scheme using the same number of parts outperforms it by $1\%$ ($84.1\% \to 85.1\%$).

\subsection{Reward Strategy Visualization}
\begin{figure}[!t]
\begin{center}
\includegraphics[scale = 0.4]{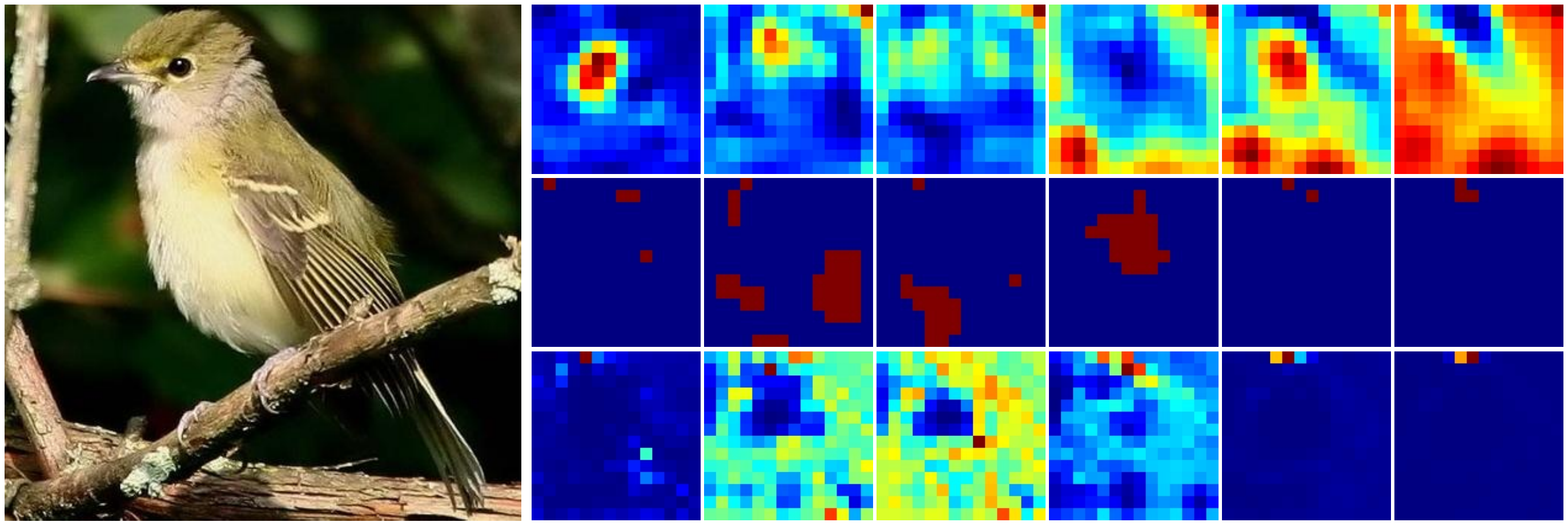}
\end{center}
\caption{Our algorithm uses 200 iterations to locate ``head'' of the bird in the left part.
 The top row in the right part shows the multi-label attribute prediction loss at different positions,
and a lighter position has higher loss.
The middle row shows the rewards at different positions. The localizer is encouraged to focus on the light position.
The bottom row shows the probability map of the localizer. Lighter positions indicate larger probability of localization.
}\label{fig:fig4}
\end{figure}

The rewards during reinforcement learning algorithm are visualized in Figure ~\ref{fig:fig4}.
From left to right, we show the heatmaps of 1st, 40-th, 80-th, 120-th, 160-th, and 200-th iterations for multi-label attribute prediction loss, rewards and the output probability of the localizer.
As can be seen, after an initial divergence on localizer probability map, the output of the localizer converges to the ``head'' position during training as expected.

\section{Conclusion}
In this paper, we present an attention localization scheme for fine-grained recognition that learns part localizers from its attribute descriptions.
An efficient reinforcement learning scheme is proposed for this task.
The proposed scheme consists of a reward function for encouraging different part localizers to capture complementary information.
It is also highly computationally efficient when the number of attributes and parts is large.
Comprehensive experiments show that our scheme obtains good part localization, improves attribute prediction, and achieves state-of-the-art recognition results on fine-grained recognition.
In the future, we will continue our efforts to improve the models of geometric part location regularization.

\medskip

\small

\bibliographystyle{./IEEEtran}
\bibliography{./IEEEabrv,./IEEEexample}

\end{document}